\documentclass{article}

\usepackage{microtype}
\usepackage{graphicx}
\usepackage{subfigure}
\usepackage{booktabs} 
\usepackage{wrapfig}

\usepackage{hyperref}

\usepackage[accepted]{icml2020}

\icmltitlerunning{Graph Structure of Neural Networks} 
\usepackage{booktabs}       
\usepackage{amsfonts}       
\usepackage{microtype}      
\usepackage{xcolor}
\usepackage{url}
\usepackage{verbatim} 
\usepackage{graphicx}
\usepackage{caption} 
\usepackage{multirow}
\usepackage{xspace}
\usepackage{amsmath}
\usepackage{amsthm}
\usepackage{amssymb}
\usepackage{times}
\usepackage{xr}
\usepackage{bbm}
\usepackage{enumitem}
\usepackage{hyperref}       
\usepackage{multicol}
\usepackage{array}
\usepackage{makecell}

\usepackage{perpage} 
\MakePerPage{footnote} 

\newcommand{\hide}[1]{}

\newcommand{\xhdr}[1]{{\noindent\bfseries #1}.}
\newcommand{\xhdrq}[1]{{\noindent\bfseries #1}}

\newcommand{\mb}{\mathbf}

\newcommand{\eg}{\emph{e.g.}}
\newcommand{\ie}{\emph{i.e.}}
\newcommand{\vs}{\emph{vs.}\xspace}
\newcommand{\threebythree}{3$\times$3}
\newcommand{\onebyone}{1$\times$1}

\newcommand{\ba}{Barab\'asi-Albert\xspace}
\newcommand{\er}{Erd\H{o}s-R\'enyi\xspace}
\newcommand{\ws}{Watts-Strogatz\xspace}

\usepackage{mathtools}

\DeclarePairedDelimiter\floor{\lfloor}{\rfloor}

\begin{document}

\twocolumn[
\icmltitle{Graph Structure of Neural Networks}



\icmlsetsymbol{equal}{*}

\begin{icmlauthorlist}
\icmlauthor{Jiaxuan You}{stanford}
\icmlauthor{Jure Leskovec}{stanford}
\icmlauthor{Kaiming He}{fair}
\icmlauthor{Saining Xie}{fair}
\end{icmlauthorlist}

\icmlaffiliation{stanford}{Department of Computer Science, Stanford University}
\icmlaffiliation{fair}{Facebook AI Research}

\icmlcorrespondingauthor{Jiaxuan You}{jiaxuan@cs.stanford.edu}
\icmlcorrespondingauthor{Saining Xie}{s9xie@fb.com}
\icmlkeywords{Machine Learning, ICML}

\vskip 0.3in
]



\printAffiliationsAndNotice{}  

\begin{abstract}
Neural networks are often represented as graphs of connections between neurons. 
However, despite their wide use, there is currently little understanding of the relationship between the graph structure of the neural network and its predictive performance.
Here we systematically investigate how does the graph structure of neural networks affect their predictive performance.
To this end, we develop a novel graph-based representation of neural networks called {\em relational graph}, 
where layers of neural network computation correspond to rounds of message exchange along the graph structure.
Using this representation we show that:
(1) a “sweet spot” of relational graphs leads to neural networks with significantly improved predictive performance; 
(2) neural network’s performance is approximately a smooth function of the clustering coefficient and average path length of its relational graph;
(3) our findings are consistent across many different tasks and datasets;
(4) the sweet spot can be identified efficiently;
(5) top-performing neural networks have graph structure surprisingly similar to those of real biological neural networks.
Our work opens new directions for the design of neural architectures and the understanding on neural networks in general.
\end{abstract}

\begin{figure*}[t]
\includegraphics[width=\linewidth]{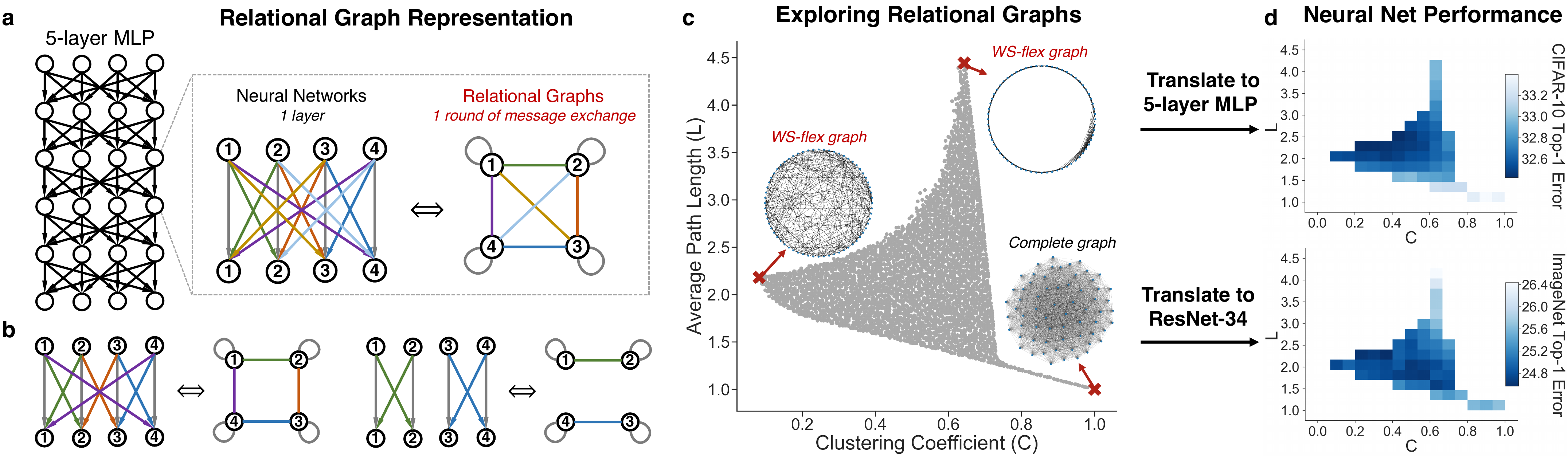}
\caption{
\textbf{Overview of our approach.} 
\textbf{(a)} A layer of a neural network can be viewed as a {\em relational graph} where we connect nodes that exchange messages. 
\textbf{(b)} More examples of neural network layers and relational graphs.
\textbf{(c)} We explore the design space of relational graphs according to their graph measures, including average path length and clustering coefficient, where the complete graph corresponds to a fully-connected layer. 
\textbf{(d)} We translate these relational graphs to neural networks and study how their predictive performance depends on the graph measures of their corresponding relational graphs.
}
\label{fig:overview}
\end{figure*}

\section{Introduction}
Deep neural networks consist of neurons organized into 
layers and connections between them.
Architecture of a neural network can be captured by its ``computational graph'' where neurons are represented as nodes and directed edges link neurons in different layers. Such graphical representation demonstrates how the network passes and transforms the information from its input neurons, through hidden layers all the way to the output neurons~\citep{mcclelland1986parallel}.

While it has been widely observed that performance of neural networks depends on their architecture~\citep{lecun1998gradient, krizhevsky2012imagenet, Simonyan2015, Szegedy2015, He2016}, there is currently little systematic understanding on the relation between a neural network's accuracy and its underlying graph structure. This is especially important for the neural architecture search, which today exhaustively searches over all possible connectivity patterns~\cite{Ying2019}.
From this perspective, several open questions arise:
Is there a systematic link between the network structure and its predictive performance? What are structural signatures of well-performing neural networks? How do such structural signatures generalize across tasks and datasets? Is there an efficient way to check whether a given neural network is promising or not?

Establishing such a relation is both scientifically and practically important because it would have direct consequences on designing more efficient and more accurate architectures. It would also inform the design of new hardware architectures that execute neural networks. Understanding the graph structures that underlie neural networks would also advance the science of deep learning.

However, establishing the relation between network architecture and its accuracy is nontrivial, because it is unclear how to map a neural network to a graph (and vice versa). The natural choice would be to use computational graph representation but it has many limitations:
(1)~\emph{lack of generality:} Computational graphs are constrained by the allowed graph properties, \eg, these graphs have to be directed and acyclic (DAGs), bipartite at the layer level, and single-in-single-out at the network level~\citep{xie2019exploring}. This limits the use of the rich tools developed for general graphs.
(2)~\emph{Disconnection with biology/neuroscience:} Biological neural networks have a much richer and less templatized structure~\citep{fornito2013graph}. There are information exchanges, rather than just single-directional flows, in the brain networks~\citep{stringer2018spontaneous}. Such biological or neurological models cannot be simply represented by directed acyclic graphs.

Here we systematically study the relationship between the graph structure of a neural network and its predictive performance. We develop a new way of representing a neural network as a graph, which we call \emph{relational graph}. Our key insight is to focus on \emph{message exchange}, rather than just on directed data flow. As a simple example, for a fixed-width fully-connected layer, we can represent one input channel \emph{and} one output channel together as a single node, and an edge in the relational graph represents the message exchange between the two nodes (Figure~\ref{fig:overview}(a)). Under this formulation, using appropriate
message exchange definition,
we show that the relational graph can represent many types of neural network layers (a fully-connected layer, a convolutional layer, 
\emph{etc.}), while getting rid of many constraints of computational graphs (such as directed, acyclic, bipartite, single-in-single-out). One neural network layer corresponds to one round of message exchange over a relational graph, and to obtain deep networks, we perform message exchange over the same graph for \emph{several rounds}.
Our new representation enables us to build neural networks that are richer and more diverse and analyze them using well-established tools of network science~\cite{barabasi2016network}.

\begin{table*}[!t]
\vspace{-1mm}
\resizebox{\textwidth}{!}{ \renewcommand{\arraystretch}{1}
\begin{tabular}[b]{llllll}
    \toprule
& \textbf{Fixed-width MLP} & \textbf{Variable-width MLP} & \textbf{ResNet-34} & \textbf{ResNet-34-sep} & \textbf{ResNet-50} \\
    \midrule
\textbf{Node feature $\mathbf{x}_i$} & \makecell[cl]{Scalar:\\ 1 dimension of data} & \makecell[cl]{Vector: multiple \\dimensions of data} & \makecell[cl]{Tensor: multiple \\ channels of data} & \makecell[cl]{Tensor: multiple \\channels of data} & \makecell[cl]{Tensor: multiple \\channels of data} \\\midrule
\textbf{Message function $f_i(\cdot)$} & Scalar multiplication & \makecell[cl]{(Non-square) \\matrix multiplication} & \threebythree{} Conv & \makecell[cl]{\threebythree{} depth-wise \\ and \onebyone{} Conv} & \makecell[cl]{\threebythree{} \\ and \onebyone{} Conv}\\ \midrule
\textbf{Aggregation function $\textsc{Agg}(\cdot)$} & $\sigma(\sum(\cdot))$ & $\sigma(\sum(\cdot))$ & $\sigma(\sum(\cdot))$ & $\sigma(\sum(\cdot))$ & $\sigma(\sum(\cdot))$  \\ \midrule
\textbf{Number of rounds $R$} & 1 round per layer & 1 round per layer & \makecell[cl]{34 rounds with\\ residual connections} &  \makecell[cl]{34 rounds with\\ residual connections} & \makecell[cl]{50 rounds with\\ residual connections} \\ 
\bottomrule
\end{tabular}}
\vspace{-5mm}
\caption{
\textbf{Diverse neural architectures expressed in the language of relational graphs.} 
These architectures are usually implemented as complete relational graphs, while we systematically explore more graph structures for these architectures.
}

\label{tb:rg_definition}
\end{table*}

We then design a graph generator named WS-flex that allows us to systematically explore the design space of neural networks (\ie, relation graphs). Based on the insights from neuroscience, we characterize neural networks by the clustering coefficient and average path length of their relational graphs (Figure~\ref{fig:overview}(c)). Furthermore, our framework is flexible and general, as we can translate relational graphs into diverse neural architectures, including Multilayer Perceptrons (MLPs), Convolutional Neural Networks (CNNs), ResNets, etc. with controlled computational budgets (Figure~\ref{fig:overview}(d)).

Using standard image classification datasets CIFAR-10 and ImageNet, we conduct a systematic study on how the architecture of neural networks
affects their predictive performance. We make several important empirical observations:
\begin{itemize}[noitemsep,topsep=0pt]
\setlength\itemsep{0.1em}
\item A ``sweet spot'' of relational graphs lead to neural networks with significantly improved  performance under controlled computational budgets (Section \ref{subsec:sweet_spot}).
\item Neural network's  performance is approximately a smooth function of the clustering coefficient and average path length of its relational graph.  (Section \ref{subsec:smooth_func}).
\item Our findings are consistent across many architectures (MLPs, CNNs, ResNets, EfficientNet) and tasks (CIFAR-10, ImageNet) (Section \ref{subsec:consistent_func}).
\item The sweet spot can be identified efficiently. Identifying a sweet spot only requires a few samples of relational graphs and a few epochs of training (Section \ref{subsec:quick_identification}).
\item Well-performing neural networks have graph structure surprisingly similar to those of real biological neural networks (Section \ref{subsec:connections}). 
\end{itemize}

Our results have implications for designing neural network architectures, advancing the science of deep learning and improving our understanding of neural networks in general\footnote{Code for experiments and analyses are available at \url{https://github.com/facebookresearch/graph2nn}}.

\section{Neural Networks as Relational Graphs}
\label{sec:rg}

To explore the graph structure of neural networks, we first introduce the concept of our \emph{relational graph} representation and its instantiations. We demonstrate how our representation can capture diverse neural network architectures under a unified framework. Using the language of graph in the context of deep learning helps bring the two worlds together and establish a foundation for our study.

\vspace{-2mm}
\subsection{Message Exchange over Graphs}
\label{subsec:graphs_message_passing}
We start by revisiting the definition of a neural network from the graph perspective.
We define a graph $G=(\mathcal{V}, \mathcal{E})$ by its node set $\mathcal{V}=\{v_1,...,v_n\}$ and edge set $\mathcal{E}\subseteq\{(v_i,v_j)|v_i,v_j\in \mathcal{V}\}$. 
We assume each node $v$ has a node feature scalar/vector $x_v$.

We call graph $G$ a \emph{relational graph}, when it is associated with message exchanges between neurons. 
Specifically, a message exchange is defined by a message function, whose input is a node's feature and output is a message, and an aggregation function, whose input is a set of messages and output is the updated node feature.
At each round of message exchange, each node sends messages to its neighbors, and aggregates incoming messages from its neighbors. Each message is transformed at each edge through a message function $f(\cdot)$, then they are aggregated at each node via an aggregation function $\textsc{Agg}(\cdot)$. 
Suppose we conduct $R$ rounds of message exchange, then the $r$-th round of message exchange for a node $v$ can be described as
\begin{equation}
\label{eq:message_passing}
    \mb{x}_v^{(r+1)} = \textsc{Agg}^{(r)}(\{f_v^{(r)}(\mb{x}_u^{(r)}),\forall u \in N(v)\})
\end{equation}
where $u, v$ are nodes in $G$, $N(v)=\{u|v \lor (u,v)\in\mathcal{E}\}$ is the neighborhood of node $v$ where we assume all nodes have self-edges, $\mb{x}_v^{(r)}$ is the input node feature and $\mb{x}_v^{(r+1)}$ is the output node feature for node $v$.
Note that this message exchange can be defined over any graph $G$; for simplicity, we only consider undirected graphs in this paper.

\begin{figure}[t]\centering
{\includegraphics[width=\linewidth]{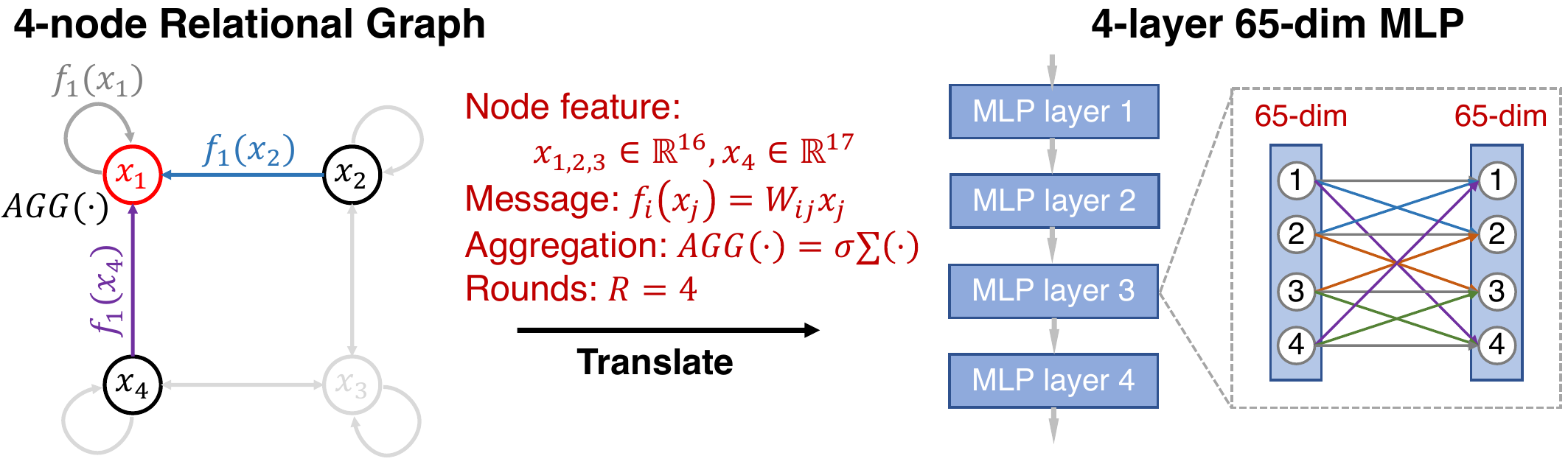}}
\caption{\textbf{Example of translating a 4-node relational graph to a 4-layer 65-dim MLP.} We highlight the message exchange for node $x_1$. Using different definitions of $x_i, f_i(\cdot), \textsc{Agg}(\cdot)$ and $R$ (those defined in Table \ref{tb:rg_definition}), relational graphs can be translated to diverse neural architectures.}
\vspace{-2mm}
\label{fig:example}
\end{figure}

Equation \ref{eq:message_passing} provides a general definition for message exchange. In the remainder of this section, we discuss how this general message exchange definition can be instantiated as different neural architectures. We summarize the different instantiations in Table \ref{tb:rg_definition}, and provide a concrete example of instantiating a 4-layer 65-dim MLP in Figure \ref{fig:example}.

\subsection{Fixed-width MLPs as Relational Graphs}
\label{subsec:graph2ffnn}
A Multilayer Perceptron (MLP) consists of layers of computation units (neurons), where each neuron performs a weighted sum over scalar inputs and outputs, followed by some non-linearity. 
Suppose the $r$-th layer of an MLP takes $\mathbf{x}^{(r)}$ as input and $\mathbf{x}^{(r+1)}$ as output, then a neuron computes:
\vspace{-2mm}
\begin{equation}
x_i^{(r+1)}=\sigma(\sum_j w_{ij}^{(r)}x_j^{(r)})
\vspace{-1mm}
\end{equation}
where $w_{ij}^{(r)}$ is the trainable weight, $x_j^{(r)}$ is the $j$-th dimension of the input $\mathbf{x}^{(r)}$, \ie, $\mathbf{x}^{(r)} = (x_1^{(r)}, ..., x_n^{(r)})$, $x_i^{(r+1)}$ is the $i$-th dimension of the output $\mathbf{x}^{(r+1)}$, and $\sigma$ is the non-linearity.

Let us consider the special case where the input and output of all the layers $x^{(r)}, 1 \leq r \leq R$ have the same feature dimensions.
In this scenario, a fully-connected, fixed-width MLP layer can be expressed with a \emph{complete relational graph}, where each node $x_i$ connects to all the other nodes $\{x_1,\dots,x_n\}$, \ie{}, neighborhood set $N(v)=\mathcal{V}$ for each node $v$.
Additionally, a fully-connected fixed-width MLP layer has a special message exchange definition, where the message function is $f_i(x_j)=w_{ij}x_i$, and the aggregation function is $\textsc{Agg}(\{x_i\})=\sigma(\sum\{x_i\})$.

The above discussion reveals that a fixed-width MLP can be viewed as a complete relational graph with a special message exchange function. Therefore, a fixed-width MLP is a special case under a much more general model family, where \textit{the message function, aggregation function, and most importantly, the relation graph structure can vary}.

This insight allows us to generalize fixed-width MLPs from using complete relational graph to any general relational graph $G$. Based on the general definition of message exchange in Equation \ref{eq:message_passing}, we have:
\vspace{-1mm}
\begin{equation}
\label{eq:mlp_graph}
x_i^{(r+1)}=\sigma(\sum_{j\in N(i)} w_{ij}^{(r)}x_j^{(r)})    
\vspace{-3mm}
\end{equation}

where $i, j$ are nodes in $G$ and $N(i)$ is defined by $G$. 

\subsection{General Neural Networks as Relational Graphs}
\label{subsec:extension}

The graph viewpoint in Equation \ref{eq:mlp_graph} lays the foundation of representing fixed-width MLPs as relational graphs. 
In this section, we discuss how we can further generalize relational graphs to general neural networks.

\xhdr{Variable-width MLPs as relational graphs}
An important design consideration for general neural networks is that layer width often varies through out the network. For example, in CNNs, a common practice is to double the layer width (number of feature channels) after spatial down-sampling.
To represent neural networks with variable layer width, we generalize node features from scalar $x_i^{(r)}$ to vector $\mathbf{x}_i^{(r)}$ which consists of some dimensions of the input $\mathbf{x}^{(r)}$ for the MLP , \ie{}, $\mathbf{x}^{(r)} = \textsc{Concat}(\mathbf{x}_1^{(r)}, ..., \mathbf{x}_n^{(r)})$, and generalize message function $f_i(\cdot)$ from scalar multiplication to matrix multiplication:
\vspace{-2mm}
\begin{equation}
\label{eq:mlp_graph_vector}
\mathbf{x}_i^{(r+1)}=\sigma(\sum_{j \in N(i)} \mathbf{W}_{ij}^{(r)}\mathbf{x}_j^{(r)})
\vspace{-2mm}
\end{equation}
where $\mathbf{W}_{ij}^{(r)}$ is the weight matrix.
Additionally, we allow that: (1) the same node in different layers, $\mathbf{x}_i^{(r)}$ and $\mathbf{x}_{i}^{(r+1)}$, can have different dimensions; (2) Within a layer, different nodes in the graph, $\mathbf{x}_i^{(r)}$ and $\mathbf{x}_{j}^{(r)}$, can have different dimensions.
This generalized definition leads to a flexible graph representation of neural networks, as we can reuse the same relational graph across different layers with arbitrary width.
Suppose we use an $n$-node relational graph to represent an $m$-dim layer, then $m\bmod n$ nodes have $\floor{m/n}+1$ dimensions each, remaining $n-(m \bmod n)$ nodes will have $\floor{m/n}$ dimensions each.
For example, if we use a 4-node relational graph to represent a 2-layer neural network. If the first layer has width of 5 while the second layer has width of 9, then the 4 nodes have dimensions $\{2,1,1,1\}$ in the first layer and $\{3,2,2,2\}$ in the second layer.

Note that under this definition, the maximum number of nodes of a relational graph is bounded by the width of the narrowest layer in the corresponding neural network (since the feature dimension for each node must be at least 1).

\xhdr{CNNs as relational graphs}
We further make relational graphs applicable to CNNs where the input becomes an image tensor $\mathbf{X}^{(r)}$. We generalize the definition of node features from vector $\mathbf{x}_i^{(r)}$ to tensor $\mathbf{X}_i^{(r)}$ that consists of some channels of the input image $\mathbf{X}^{(r)} = \textsc{Concat}(\mathbf{X}_1^{(r)}, ..., \mathbf{X}_n^{(r)})$. We then generalize the message exchange definition with convolutional operator. Specifically,
\vspace{-2mm}
\begin{equation}
\label{eq:mlp_graph_vector}
\mathbf{X}_i^{(r+1)}=\sigma(\sum_{j \in N(i)} \mathbf{W}_{ij}^{(r)} * \mathbf{X}_j^{(r)})
\vspace{-2mm}
\end{equation}
where $*$ is the convolutional operator and $\mathbf{W}_{ij}^{(r)}$ is the convolutional filter.
Under this definition, the widely used dense convolutions are again represented as complete graphs.

\xhdr{Modern neural architectures as relational graphs}
Finally, we generalize relational graphs to represent modern neural architectures with more sophisticated designs. For example, to represent a ResNet~\citep{He2016}, we keep the residual connections between layers unchanged. To represent neural networks with bottleneck transform~\cite{He2016}, a relational graph alternatively applies message exchange with \threebythree{} and \onebyone{} convolution; similarly, in the efficient computing setup, the widely used separable convolution~\cite{howard2017mobilenets, chollet2017xception} can be viewed as alternatively applying message exchange with \threebythree{} depth-wise convolution and \onebyone{} convolution.

Overall, relational graphs provide a general representation for neural networks. With proper definitions of node features and message exchange, relational graphs can represent diverse neural architectures, as is summarized in Table \ref{tb:rg_definition}.

\section{Exploring Relational Graphs}
\label{sec:erg}
In this section, we describe in detail how we design and explore the space of relational graphs defined in Section \ref{sec:rg}, in order to study the relationship between the graph structure of neural networks and their predictive performance.
Three main components are needed to make progress: (1) \emph{graph measures} that characterize graph structural properties, (2) \emph{graph generators} that can generate diverse graphs, and (3) a way to \emph{control the computational budget}, so that the differences in performance of different neural networks are due to their diverse relational graph structures.

\subsection{Selection of Graph Measures}
\label{subsec:graph_stats}

Given the complex nature of graph structure, \emph{graph measures} are often used to characterize graphs.
In this paper, we focus on one global graph measure, average path length, and one local graph measure, clustering coefficient.
Notably, these two measures are widely used in network science~\citep{Watts1998} and neuroscience~\citep{sporns2003graph, bassett2006small}.
Specifically, average path length measures the average shortest path distance between any pair of nodes; clustering coefficient measures the proportion of edges between the nodes within a given node's neighborhood, divided by the number of edges that could possibly exist between them, averaged over all the nodes.
There are other graph measures that can be used for analysis, which are included in the Appendix.

\subsection{Design of Graph Generators}
\label{subsec:generators}

Given selected graph measures,
we aim to generate diverse graphs that can cover a large span of graph measures, using a \emph{graph generator}. 
However, such a goal requires careful generator designs: classic graph generators can only generate a limited class of graphs, while recent learning-based graph generators are designed to imitate given exemplar graphs~\citep{kipf2016semi,li2018learning,you2018graph,you2018graphrnn,you2019g2sat}.

\xhdr{Limitations of existing graph generators}
To illustrate the limitation of existing graph generators, we investigate the following classic graph generators:
(1) \er (ER) model that can sample graphs with given node and edge number uniformly at random~\citep{Erdos1960}; (2) \ws(WS) model that can generate graphs with small-world properties~\citep{Watts1998}; (3) \ba (BA) model that can generate scale-free graphs~\citep{Albert2002}; (4) Harary model that can generate graphs with maximum connectivity~\citep{harary1962maximum}; (5) regular ring lattice graphs (ring graphs); (6) complete graphs. 
For all types of graph generators, we control the number of nodes to be 64, enumerate all possible discrete parameters and grid search over all continuous parameters of the graph generator. We generate 30 random graphs with different random seeds under each parameter setting. In total, we generate 
486,000 WS graphs, 53,000 ER graphs, 8,000 BA graphs, 1,800 Harary graphs, 54 ring graphs and 1 complete graph (more details provided in the Appendix).
In Figure \ref{fig:graph_stats}, we can observe that graphs generated by those classic graph generators have a limited span in the space of average path length and clustering coefficient.

\xhdr{WS-flex graph generator}
Here we propose the WS-flex graph generator that can generate graphs with a wide coverage of graph measures; notably, WS-flex graphs almost encompass all the graphs generated by classic random generators mentioned above, as is shown in Figure \ref{fig:graph_stats}.
WS-flex generator generalizes WS model by relaxing the constraint that all the nodes have the same degree before random rewiring.
Specifically, WS-flex generator is parametrized by node $n$, average degree $k$ and rewiring probability $p$. The number of edges is determined as $e=\floor{n*k/2}$. Specifically, WS-flex generator first creates a ring graph where each node connects to $\floor{e/n}$ neighboring nodes; then the generator randomly picks $e \bmod n$ nodes and connects each node to one closest neighboring node; finally, all the edges are randomly rewired with probability $p$.
We use WS-flex generator to smoothly sample within the space of clustering coefficient and average path length, then sub-sample 3942 graphs for our experiments, as is shown in Figure \ref{fig:overview}(c).

\subsection{Controlling Computational Budget}
\vspace{-1mm}
\label{subsec:computational_budget}
To compare the neural networks translated by these diverse graphs, it is important to ensure that all networks have approximately the same complexity, so that the differences in performance are due to their relational graph structures. We use FLOPS 
(\# of multiply-adds)
as the metric. We first compute the FLOPS of our baseline network instantiations (\ie{} complete relational graph), and use them as the reference complexity in each experiment. As described in Section \ref{subsec:extension}, a relational graph structure can be instantiated as a neural network with variable width, by partitioning dimensions or channels into disjoint set of node features. Therefore, we can conveniently adjust the width of a neural network to match the reference complexity (within 0.5\% of baseline FLOPS) without changing the relational graph structures. We provide more details in the Appendix.

\begin{figure}[!t]\centering
{
\vspace{-1mm}
\includegraphics[width=\linewidth]{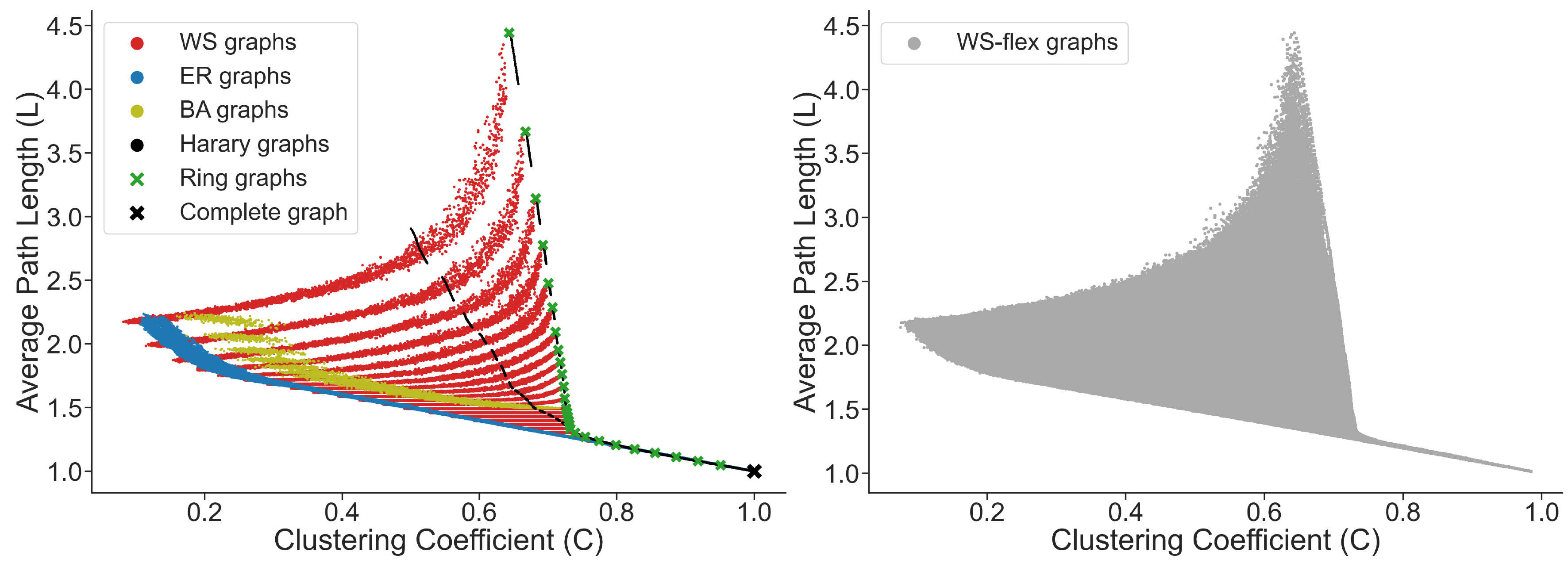}}
\vspace{-7mm}
\caption{\textbf{Graphs generated by different graph generators.} The proposed graph generator WS-flex can cover a much larger region of graph design space. WS (\ws), BA (\ba), ER (\er).}
\vspace{-1mm}
\label{fig:graph_stats}
\end{figure}

\begin{figure*}[t]
\vspace{-2mm}
\includegraphics[width=\linewidth]{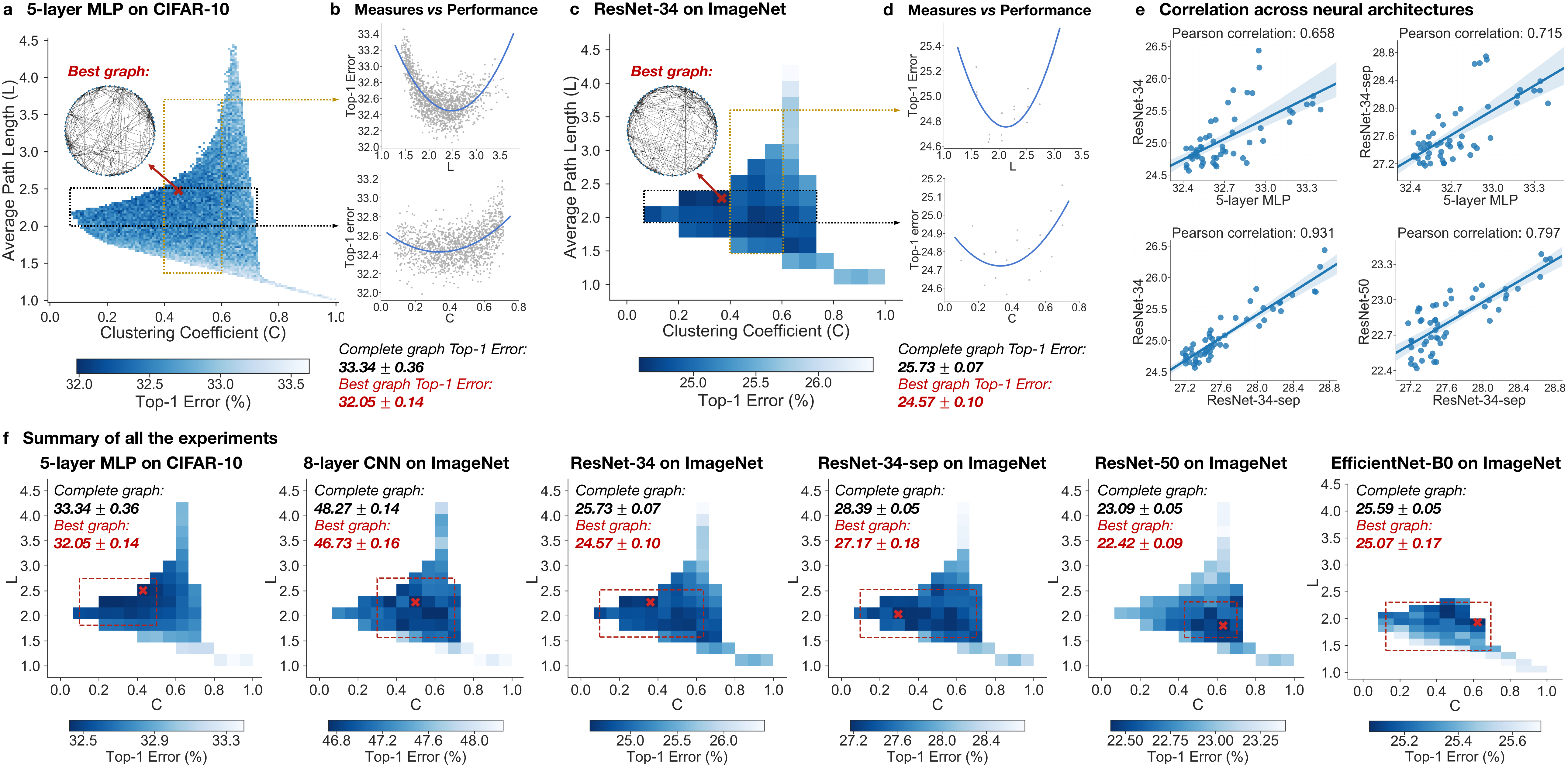}
\caption{
\textbf{Key results.} The computational budgets of all the experiments are rigorously controlled. Each visualized result is averaged over at least 3 random seeds.
A complete graph with $C=1$ and $L=1$ (lower right corner) is regarded as the baseline.
\textbf{(a)(c)} \textbf{Graph measures \vs neural network performance.} The best graphs significantly outperform the baseline complete graphs.
\textbf{(b)(d)} \textbf{Single graph measure \vs neural network performance.} Relational graphs that fall within the given range are shown as grey points. The overall smooth function is indicated by the blue regression line.
\textbf{(e)} \textbf{Consistency across architectures.} Correlations of the performance of the same set of 52 relational graphs when translated to different neural architectures are shown.
\textbf{(f)} \textbf{Summary of all the experiments.} Best relational graphs (the red crosses) consistently outperform the baseline complete graphs across different settings. Moreover, we highlight the ``sweet spots'' (red rectangular regions), in which relational graphs are not statistically worse than the best relational graphs (bins with red crosses). Bin values of 5-layer MLP on CIFAR-10 are average over all the relational graphs whose $C$ and $L$ fall into the given bin.
}
\label{fig:findings}
\vspace{-1mm}
\end{figure*}
\section{Experimental Setup}
Considering the large number of candidate graphs (3942 in total) that we want to explore, we first investigate graph structure of MLPs on the CIFAR-10 dataset~\citep{Krizhevsky09learningmultiple} which has 50K training images and 10K validation images.
We then further study the larger and more complex task of ImageNet classification~\citep{Russakovsky2015}, which consists of 1K image classes, 1.28M training images and 50K validation images.

\subsection{Base Architectures}
For CIFAR-10 experiments, We use a 5-layer MLP with 512 hidden units as the baseline architecture. The input of the MLP is a 3072-d flattened vector of the ($32\times32\times3$) image, the output is a 10-d prediction.
Each MLP layer has a ReLU non-linearity and a BatchNorm layer~\citep{ioffe2015batch}. We train the model for 200 epochs with batch size 128, using cosine learning rate schedule~\citep{loshchilov2016sgdr} with an initial learning rate of 0.1 (annealed to 0, no restarting). We train all MLP models with 5 different random seeds and report the averaged results.

For ImageNet experiments, we use three ResNet-family architectures, including (1) ResNet-34, which only consists of basic blocks of \threebythree{} convolutions~\citep{He2016}; (2) ResNet-34-sep, a variant where we replace all \threebythree{} dense convolutions in ResNet-34 with \threebythree{} separable convolutions~\citep{chollet2017xception}; (3) ResNet-50, which consists of bottleneck blocks~\citep{He2016} of \onebyone{}, \threebythree{}, \onebyone{} convolutions.
Additionally, we use EfficientNet-B0 architecture~\citep{tan2019efficientnet} that achieves good performance in the small computation regime. 
Finally, we use a simple 8-layer CNN with \threebythree{} convolutions. The model has 3 stages with [64, 128, 256] hidden units. Stride-2 convolutions are used for down-sampling. The stem and head layers are the same as a ResNet.
We train all the ImageNet models for 100 epochs using cosine learning rate schedule with initial learning rate of 0.1.
Batch size is 256 for ResNet-family models and 512 for EfficientNet-B0. We train all ImageNet models with 3 random seeds and report the averaged performance. 
All the baseline architectures have a complete relational graph structure. The reference computational complexity is 2.89e6 FLOPS for MLP, 3.66e9 FLOPS for ResNet-34, 0.55e9 FLOPS for ResNet-34-sep, 4.09e9 FLOPS for ResNet-50, 0.39e9 FLOPS for EffcientNet-B0, and 0.17e9 FLOPS for 8-layer CNN. Training an MLP model roughly takes 5 minutes on a NVIDIA Tesla V100 GPU, and training a ResNet model on ImageNet roughly takes a day on 8 Tesla V100 GPUs with data parallelism. We provide more details in Appendix.

\subsection{Exploration with Relational Graphs}
For all the architectures, we instantiate each sampled relational graph as a neural network, using the corresponding definitions outlined in Table \ref{tb:rg_definition}. Specifically, we replace all the dense layers (linear layers, \threebythree{} and \onebyone{} convolution layers) with their relational graph counterparts. We leave the input and output layer unchanged and keep all the other designs (such as down-sampling, skip-connections, \emph{etc.}) intact. We then match the reference computational complexity for all the models, as discussed in Section \ref{subsec:computational_budget}.

For CIFAR-10 MLP experiments, we study 3942 sampled relational graphs of 64 nodes as described in Section \ref{subsec:generators}.
For ImageNet experiments, due to high computational cost, we sub-sample 52 graphs uniformly from the 3942 graphs. 
Since EfficientNet-B0 is a small model with a layer that has only 16 channels, we can not reuse the 64-node graphs sampled for other setups. We re-sample 48 relational graphs with 16 nodes following the same procedure in Section \ref{sec:erg}.

\section{Results}
\label{sec:results}
In this section, we summarize the results of our experiments and discuss our key findings.
We collect top-1 errors for all the sampled relational graphs on different tasks and architectures, and also record the graph measures (average path length $L$ and clustering coefficient $C$) for each sampled graph. We present these results as heat maps of graph measures \vs predictive performance (Figure \ref{fig:findings}(a)(c)(f)).

\subsection{A Sweet Spot for Top Neural Networks}
\label{subsec:sweet_spot}

Overall, the heat maps of graph measures \vs predictive performance (Figure \ref{fig:findings}(f)) show that there \emph{exist} graph structures that can outperform the complete graph (the pixel on bottom right) baselines. The best performing relational graph can outperform the complete graph baseline by 1.4\% top-1 error on CIFAR-10, and 0.5\% to 1.2\% for models on ImageNet. 

Notably, we discover that top-performing graphs tend to cluster into a sweet spot in the space defined by $C$ and $L$ (red rectangles in Figure \ref{fig:findings}(f)).
We follow these steps to identify a sweet spot: (1) we downsample and aggregate the 3942 graphs in Figure \ref{fig:findings}(a) into a coarse resolution of 52 bins, where each bin records the performance of graphs that fall into the bin;
(2) we identify the bin with best average performance (red cross in Figure \ref{fig:findings}(f));
(3) we conduct one-tailed \emph{t}-test over each bin against the best-performing bin, and record the bins that are \emph{not} significantly worse than the best-performing bin (\emph{p}-value 0.05 as threshold). The minimum area rectangle that covers these bins is visualized as a sweet spot. For 5-layer MLP on CIFAR-10, the sweet spot is $C\in [0.10, 0.50]$, $L\in[1.82,2.75]$.

\subsection{Neural Network Performance as a Smooth Function over Graph Measures}
\label{subsec:smooth_func}

In Figure \ref{fig:findings}(f), we observe that neural network's predictive performance is approximately a smooth function of the clustering coefficient and average path length of its relational graph. 
Keeping one graph measure fixed in a small range ($C\in[0.4,0.6]$, $L\in[2,2.5]$), we visualize network performances against the other measure (shown in Figure \ref{fig:findings}(b)(d)).
We use second degree polynomial regression to visualize the overall trend. We observe that both clustering coefficient and average path length are indicative of neural network performance, demonstrating a smooth U-shape correlation.

\subsection{Consistency across Architectures} 
\label{subsec:consistent_func}
Given that relational graph defines a shared design space across various neural architectures, we observe that relational graphs with certain graph measures may consistently perform well regardless of how they are instantiated.

\xhdr{Qualitative consistency}
We visually observe in Figure \ref{fig:findings}(f) that the sweet spots are roughly consistent across different architectures.
Specifically, if we take the union of the sweet spots across architectures, we have $C \in [0.43, 0.50]$, $L \in [1.82, 2.28]$ which is the consistent sweet spot across architectures.
Moreover, the U-shape trends between graph measures and corresponding neural network performance, shown in Figure \ref{fig:findings}(b)(d), are also visually consistent.

\xhdr{Quantitative consistency}
To further quantify this consistency across tasks and architectures, we select the 52 bins in the heat map in Figure \ref{fig:findings}(f), where the bin value indicates the average performance of relational graphs whose graph measures fall into the bin range. We plot the correlation of the 52 bin values across different pairs of tasks, shown in Figure \ref{fig:findings}(e).
We observe that the performance of relational graphs with certain graph measures correlates across different tasks and architectures. For example, even though a ResNet-34 has much higher complexity than a 5-layer MLP, and ImageNet is a much more challenging dataset than CIFAR-10, a fixed set relational graphs would perform similarly in both settings, indicated by a Pearson correlation of 0.658 (p-value $ < 10^{-8}$).

\subsection{Quickly Identifying a Sweet Spot}
\label{subsec:quick_identification}

\begin{figure}[!tp]\centering
{
\includegraphics[width=\linewidth]{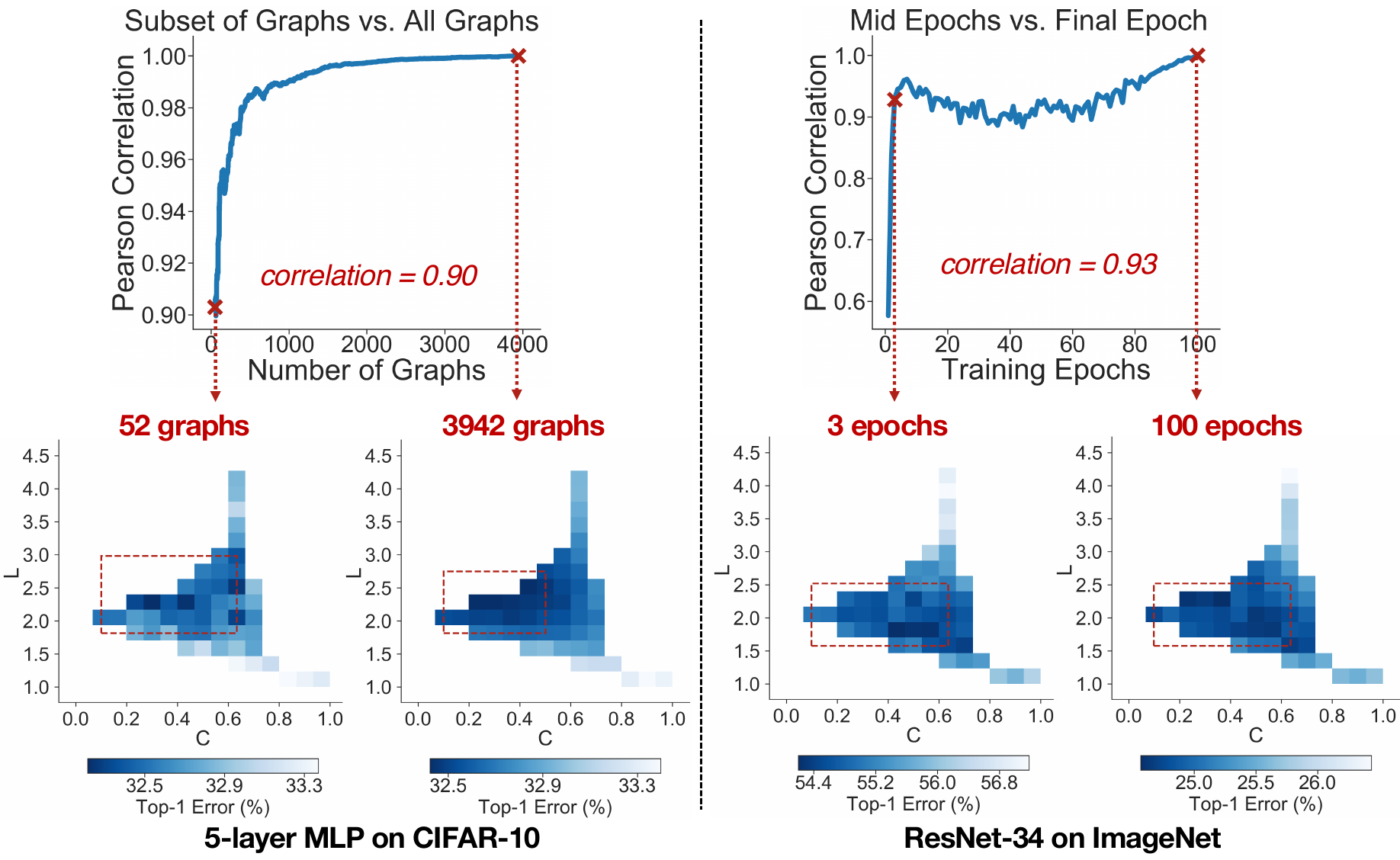}}
\caption{
\textbf{Quickly identifying a sweet spot.}
\textbf{Left:} The correlation between sweet spots identified using fewer samples of relational graphs and using all 3942 graphs.
\textbf{Right:} The correlation between sweet spots identified at the intermediate training epochs and the final epoch (100 epochs).}
\label{fig:corr_epochgraph}
\end{figure}

Training thousands of relational graphs until convergence might be computationally prohibitive. Therefore, we quantitatively show that a sweet spot can be identified with much less computational cost, \eg{}, by sampling fewer graphs and training for fewer epochs.

\xhdrq{How many graphs are needed?}
Using the 5-layer MLP on CIFAR-10 as an example, we consider the heat map over 52 bins in Figure \ref{fig:findings}(f) which is computed using 3942 graph samples.
We investigate if a similar heat map can be produced with much fewer graph samples.
Specifically, we sub-sample the graphs in each bin while making sure each bin has at least one graph. We then compute the correlation between the 52 bin values computed using all 3942 graphs and using sub-sampled fewer graphs, as is shown in Figure \ref{fig:corr_epochgraph} (left).
We can see that bin values computed using only 52 samples have a high $0.90$ Pearson correlation with the bin values computed using full 3942 graph samples. This finding suggests that, in practice, much fewer graphs are needed to conduct a similar analysis.

\xhdrq{How many training epochs are needed?}
Using ResNet-34 on ImageNet as an example, we compute the correlation between the validation top-1 error of partially trained models and the validation top-1 error of models trained for full 100 epochs, over the 52 sampled relational graphs, as is visualized in Figure \ref{fig:corr_epochgraph} (right).
Surprisingly, models trained after 3 epochs already have 
a high correlation (0.93), which means that \emph{good relational graphs perform well even at the early training epochs}. This finding is also important as it indicates that the computation cost to determine if a relational graph is promising can be greatly reduced.

\begin{table}[!t]
\centering
\resizebox{\columnwidth}{!}{ \renewcommand{\arraystretch}{1}
\begin{tabular}[b]{p{4cm}ccc}
\toprule
\textbf{Graph}  & \textbf{Path (L)} & \textbf{Clustering (C)} & \textbf{CIFAR-10 Error (\%)} \\ \midrule
Complete graph & 1.00 & 1.00 & 33.34 $\pm$ 0.36 \\ \midrule
Cat cortex  & 1.81 & 0.55 & 33.01 $\pm$ 0.22 \\ \midrule
Macaque visual cortex  & 1.73 & 0.53 & 32.78 $\pm$ 0.21 \\ \midrule
\textbf{Macaque whole cortex}  & \textbf{2.38} & \textbf{0.46} & 32.77 $\pm$ 0.14 \\ 
\midrule
\parbox{4cm}{\textbf{Consistent sweet spot\\across neural architectures}} & \textbf{1.82-2.28} & \textbf{0.43-0.50} & 32.50 $\pm$ 0.33 \\
\midrule
\textbf{Best 5-layer MLP} & \textbf{2.48} & \textbf{0.45} & 32.05 $\pm$ 0.14 \\
\bottomrule
\end{tabular}}
\vspace{-6mm}
\caption{Top artificial neural networks can be similar to biological neural networks ~\citep{bassett2006small}. }
\label{tb:brain_stats}
\end{table}

\begin{figure}[!tp]\centering
{
\includegraphics[width=\linewidth]{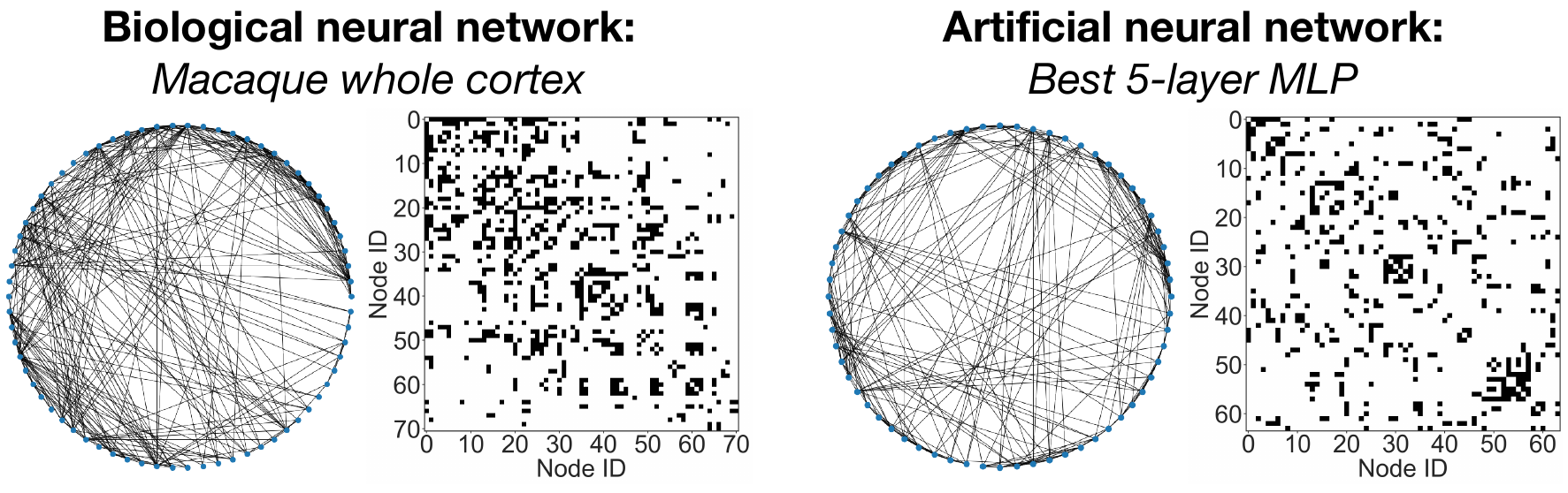}}
\vspace{-6mm}
\caption{Visualizations of graph structure of biological (\textbf{left}) and artificial (\textbf{right}) neural networks.}
\label{fig:bio}
\end{figure}

\subsection{Network Science and Neuroscience Connections}
\label{subsec:connections}
\xhdr{Network science}
The average path length that we measure characterizes how well information is exchanged across the network~\citep{latora2001efficient}, which aligns with our definition of relational graph that consists of rounds of message exchange.
Therefore, the U-shape correlation in Figure \ref{fig:findings}(b)(d) might indicate a trade-off between message exchange efficiency \cite{sengupta2013information} and capability of learning \emph{distributed representations} \cite{hinton1984distributed}.

\xhdr{Neuroscience}
The best-performing relational graph that we discover surprisingly resembles biological neural networks, as is shown in Table \ref{tb:brain_stats} and Figure \ref{fig:bio}. 
The similarities are in two-fold: (1) the graph measures ($L$ and $C$) of top artificial neural networks are highly similar to biological neural networks; (2) with the relational graph representation, we can translate biological neural networks to 5-layer MLPs, and found that these networks also outperform the baseline complete graphs. While our findings are preliminary, our approach opens up new possibilities for interdisciplinary research in network science, neuroscience and deep learning.

\section{Related Work}
\xhdr{Neural network connectivity} The design of neural network connectivity patterns has been focused on \emph{computational graphs} at different granularity: the macro structures, \ie{} connectivity across layers~\citep{lecun1998gradient, krizhevsky2012imagenet, Simonyan2015, Szegedy2015, He2016, Huang2017, tan2019efficientnet}, and the micro structures, \ie{} connectivity within a layer~\citep{lecun1998gradient, Xie2017, Zhang2018, howard2017mobilenets, dao2019learning, alizadeh2019butterfly}. Our current exploration focuses on the latter, but the same methodology can be extended to the macro space. Deep Expander Networks \citep{prabhu2018deep} adopt expander graphs to generate bipartite structures. RandWire \citep{xie2019exploring} generates macro structures using existing graph generators. However, the statistical relationships between graph structure measures and network predictive performances were not explored in those works. Another related work is Cross-channel Communication Networks~\citep{Yang2019Cross} which aims to encourage the neuron communication through message passing, where only a complete graph structure is considered.

\xhdr{Neural architecture search} Efforts on learning the connectivity patterns at micro \citep{ahmed2018maskconnect, wortsman2019discovering, yang2018convolutional}, or macro~\citep{Zoph2017, Zoph2018} level mostly focus on improving learning/search algorithms~\citep{Liu2018a, Pham2018, Real2018, Liu2019}. NAS-Bench-101~\citep{Ying2019} defines a graph search space by enumerating DAGs with constrained sizes ($\leq7$ nodes, \emph{cf.} 64-node graphs in our work). Our work points to a new path: instead of exhaustively searching over all the possible connectivity patterns, certain graph generators and graph measures could define a smooth space where the search cost could be significantly reduced.

\section{Discussions}

\xhdr{Hierarchical graph structure of neural networks} 
As the first step in this direction, our work focuses on graph structures at the layer level. Neural networks are intrinsically hierarchical graphs (from connectivity of neurons to that of layers, blocks, and networks) which constitute a more complex design space than what is considered in this paper. Extensive exploration in that space will be computationally prohibitive, but we expect our methodology and findings to generalize.

\xhdr{Efficient implementation}
Our current implementation uses standard CUDA kernels thus relies on weight masking, which leads to worse wall-clock time performance compared with baseline complete graphs. However, the practical adoption of our discoveries is not far-fetched. Complementary to our work, there are ongoing efforts such as block-sparse kernels~\citep{gray2017gpu} and fast sparse ConvNets~\citep{elsen2019fast} which could close the gap between theoretical FLOPS and real-world gains. Our work might also inform the design of new hardware architectures, \eg, biologically-inspired ones with spike patterns~\citep{pei2019towards}.

\xhdr{Prior \emph{vs.} Learning} 
We currently utilize the relational graph representation as a \emph{structural prior}, \ie, we hard-wire the graph structure on neural networks throughout training. It has been shown that deep ReLU neural networks can automatically learn sparse representations~\citep{glorot2011deep}. A further question arises: without imposing graph priors, does any graph structure emerge from training a (fully-connected) neural network?

\begin{figure}[!h]\centering
{
\includegraphics[width=0.65\linewidth]{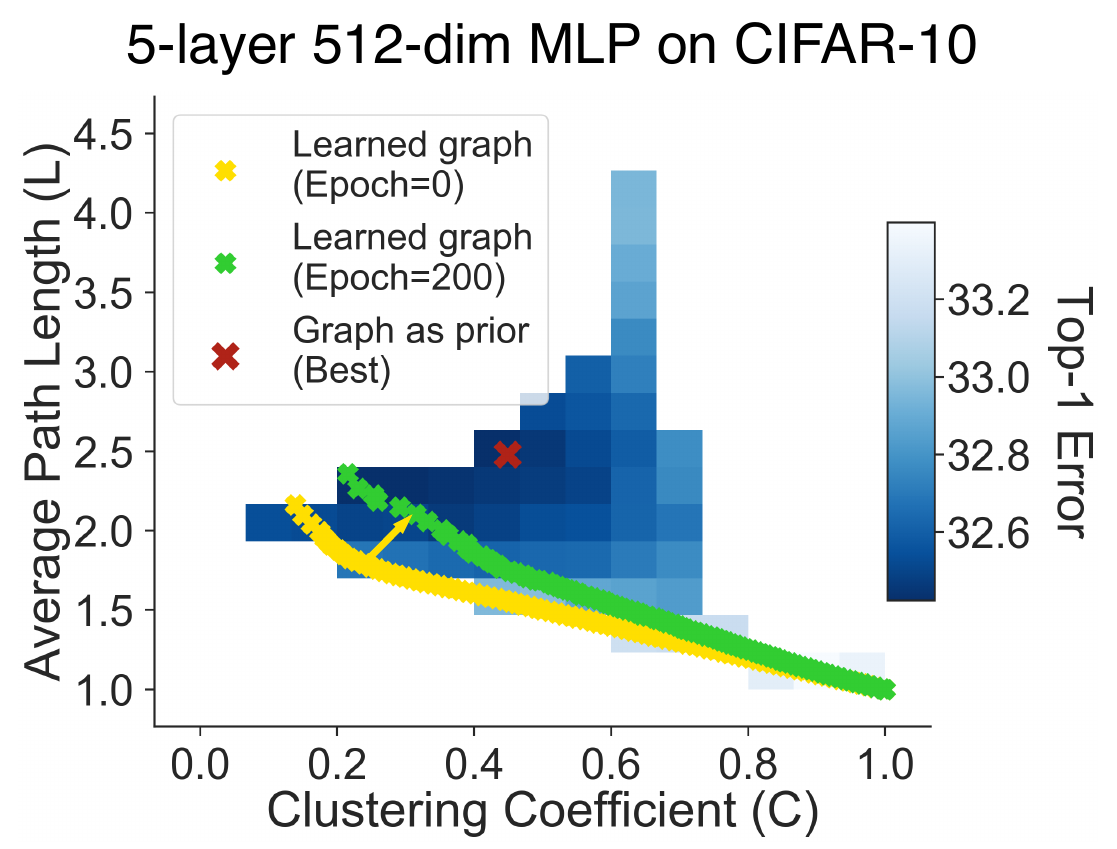}}
\caption{
\textbf{Prior \emph{vs.} Learning.}
Results for 5-layer MLPs on CIFAR-10. We highlight the best-performing graph when used as a \emph{structural prior}. Additionally, we train a fully-connected MLP, and visualize the learned weights as a relational graph (different points are graphs under different thresholds). The learned graph structure moves towards the ``sweet spot'' after training but does not close the gap.
}
\label{fig:discussion}
\end{figure}

As a preliminary exploration, we ``reverse-engineer'' a trained neural network and study the emerged relational graph structure. 
Specifically, we train a fully-connected 5-layer MLP on CIFAR-10 (the same setup as in previous experiments).
We then try to infer the underlying relational graph structure of the network via the following steps: (1) to get nodes in a relational graph, we stack the weights from all the hidden layers and group them into 64 nodes, following the procedure described in Section \ref{subsec:graph2ffnn}; (2) to get undirected edges, the weights are summed by their transposes; (3) we compute the Frobenius norm of the weights as the edge value; (4) we get a sparse graph structure by binarizing edge values with a certain threshold. 

We show the extracted graphs under different thresholds in Figure \ref{fig:discussion}. As expected, the extracted graphs at initialization follow the patterns of E-R graphs (Figure \ref{fig:graph_stats}(left)), since weight matrices are randomly i.i.d. initialized.
Interestingly, after training to convergence, the extracted graphs are no longer E-R random graphs and move towards the sweet spot region we found in Section \ref{sec:results}. Note that there is still a gap between these learned graphs and the best-performing graph imposed as a structural prior, which might explain why a fully-connected MLP has inferior performance.

In our experiments, we also find that there are a few special cases where learning the graph structure can be superior (\ie, when the task is simple and the network capacity is abundant). We provide more discussions in the Appendix. Overall, these results further demonstrate that studying the graph structure of a neural network is crucial for understanding its predictive performance.

\xhdr{Unified view of Graph Neural Networks (GNNs) and general neural architectures} The way we define neural networks as a message exchange function over graphs is partly inspired by GNNs~\citep{kipf2016semi, hamilton2017inductive, velivckovic2017graph}. 
Under the relational graph representation, we point out that \emph{GNNs are a special class of general neural architectures} where: (1) graph structure is regarded as the input instead of part of the neural architecture; consequently, (2) message functions are shared across all the edges to respect the invariance properties of the input graph.
Concretely, recall how we define general neural networks as relational graphs:
\vspace{-1mm}
\begin{equation}
\label{eq:gnn}
    \mb{x}_v^{(r+1)} = \textsc{Agg}^{(r)}(\{f_v^{(r)}(\mb{x}_u^{(r)}),\forall u \in N(v)\})
\vspace{-1mm}
\end{equation}
For a GNN, $f_v^{(r)}$ is shared across all the edges $(u,v)$, and $N(v)$ is provided by the input graph dataset, while a general neural architecture does not have such constraints.

Therefore, our work offers a \emph{unified view of GNNs and general neural architecture design}, which we hope can bridge the two communities and inspire new innovations. On one hand, successful techniques in general neural architectures can be naturally introduced to the design of GNNs, such as separable convolution \cite{howard2017mobilenets}, group normalization \cite{wu2018group} and Squeeze-and-Excitation block \cite{hu2018squeeze}; on the other hand, novel GNN architectures \cite{you2019position,chen2019equivalence} beyond the commonly used paradigm (\ie, Equation \ref{eq:gnn}) may inspire more advanced neural architecture designs.

\section{Conclusion}
In sum, 
we propose a new perspective of using relational graph representation for analyzing and understanding neural networks.
Our work suggests a new transition from studying conventional computation architecture to studying \emph{graph structure} of neural networks.
We show that well-established graph techniques and methodologies offered in other science disciplines (network science, neuroscience, \emph{etc.}) could contribute to understanding and designing deep neural networks. We believe this could be a fruitful avenue of future research that tackles more complex situations.

\section*{Acknowledgments}
This work is done during Jiaxuan You's internship at Facebook AI Research.
Jure Leskovec is a Chan Zuckerberg Biohub investigator.
The authors thank Alexander Kirillov, Ross Girshick, Jonathan Gomes Selman, Pan Li for their helpful discussions.

\bibliography{bibli}
\bibliographystyle{icml2020}


\end{document}


\twocolumn[
\icmltitle{Appendix for \\ ``Graph Structure of Neural Networks''}



\icmlsetsymbol{equal}{*}

\begin{icmlauthorlist}
\icmlauthor{Jiaxuan You}{stanford}
\icmlauthor{Jure Leskovec}{stanford}
\icmlauthor{Kaiming He}{fair}
\icmlauthor{Saining Xie}{fair}
\end{icmlauthorlist}

\icmlaffiliation{stanford}{Department of Computer Science, Stanford University}
\icmlaffiliation{fair}{Facebook AI Research}

\icmlcorrespondingauthor{Jiaxuan You}{jiaxuan@cs.stanford.edu}
\icmlcorrespondingauthor{Saining Xie}{s9xie@fb.com}

\icmlkeywords{Machine Learning, ICML}

\vskip 0.3in
]



\printAffiliationsAndNotice{}  



\newpage

\section{Details for Generating Relational Graphs}

Here we provide more details for how we generate graphs in Section 3.2.
For all generators, we fix the number of nodes $n=64$, and constrain the graph sparsity within $[0.125,1.0]$.

\xhdr{\ws (WS) graphs}
WS graphs are characterized by: (1) number of nodes $n$, (2) initial node degree $k$ (must be an integer), (3) edge rewiring probability (randomness) $p$.
We search over:
\begin{itemize}[noitemsep,topsep=0pt]
    \item degree $k\in$ \texttt{np.arange(8,62)}
    \item randomness $p\in$ \texttt{np.linspace(0,1,300)**2}
    \item 30 random seeds
\end{itemize}
Since graph measures are more sensitive when $p$ is small, we increase the sample density of small $p$ value by squaring $p$.
In total, we generate $54\times300\times30=486,000$ WS graphs.

\xhdr{\er (ER) graphs}
ER graphs are characterized by: (1) number of nodes $n$, (2) number of edges $m$.
We search over:
\begin{itemize}[noitemsep,topsep=0pt]
    \item edge number $m\in$ \texttt{np.arange($64\times4,64\times63/2$)}
    \item 30 random seeds
\end{itemize}
In total, we generate $1760\times30=52,800$ ER graphs.

\xhdr{\ba (BA) graphs}
ER graphs are characterized by: (1) number of nodes $n$, (2) number of existing nodes $m$ that a new node connects to.
We search over:
\begin{itemize}[noitemsep,topsep=0pt]
    \item $m\in$ \texttt{np.arange(4,30)}
    \item 300 random seeds
\end{itemize}
In total, we generate $26\times300=7,800$ ER graphs.

\xhdr{Harary graphs}
Harary graphs are determined by: (1) number of nodes $n$, (2) number of edges $m$.
We search over:
\begin{itemize}[noitemsep,topsep=0pt]
    \item edge number $m\in$ \texttt{np.arange(64*4,64*63/2)}
\end{itemize}
In total, we generate $1760$ Harary graphs.

\xhdr{Ring graphs}
Ring graphs are characterized by: (1) number of nodes $n$, (2) node degree $k$ (integer).
We search over:
\begin{itemize}[noitemsep,topsep=0pt]
    \item degree $k\in$ \texttt{np.arange(8,62)}
\end{itemize}
In total, we generate $54$ ring graphs.

\xhdr{WS-flex graphs}
We describe the detailed procedures of getting 3942 WS-flex graphs that we used in the experiments.
WS-flex graphs are characterized by: (1) number of nodes $n$, (2) average node degree $k$ (real number), (3) edge rewiring probability (randomness) $p$.
We search over:
\begin{itemize}[noitemsep,topsep=0pt]
    \item degree $k\in$ \texttt{np.linspace(8,62,300)}
    \item randomness $p\in$ \texttt{np.linspace(0,1,300)**2}
    \item 30 random seeds
\end{itemize}
In total, we generate $300\times300\times30=2,700,000$ WS-flex graphs. Generating these WS-flex graphs (and computing their average path length and clustering coefficient) only takes about 1 hour on a 80 CPU core machine.

Next, we sub-sample 3942 graphs from these 2.7M candidate graphs.
We create 2-d bins over the graph structure measures: (1) for average path length, we create $15\times9$ bins whose bin edges are given by \texttt{np.linspace($1,4.5,15\times9+1$)}; (2) for clustering coefficient, we create $15\times9$ bins whose bin edges are given by \texttt{np.linspace($0,1,15\times9+1$)}.
We sub-sample 1 graph, whose graph structure measures fall within a given 2-d bin, for each of the 2-d bins. After gathering bins that have graphs, we get 3942 graphs in total.

For ImageNet experiments, we further sub-sample 52 graphs from these 3942 graphs.
Specifically, we collect graphs in the bins whose bin ID $(i \mod 9) = 5$, so that the sub-sampled graphs are roughly uniformly distributed in the graph measure space.

\begin{figure*}[t]
\includegraphics[width=\linewidth]{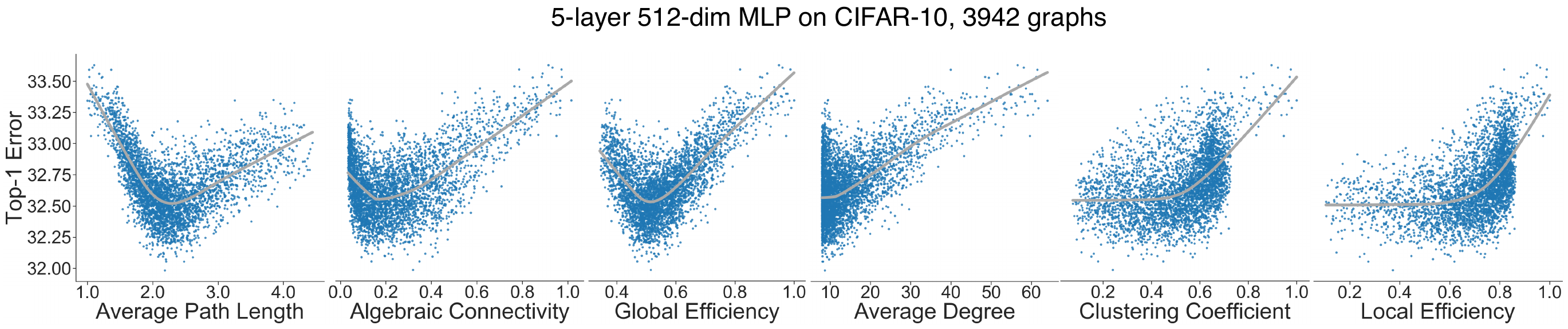}
\caption{\textbf{More graph measures \vs neural network performance.} 
Global (\textbf{left 3}) and local (\textbf{right 3}) graph measures versus 5-layer 512-dim MLP performance on CIFAR-10.}
\label{fig:1d_stats}
\end{figure*}

\begin{figure*}[!t]
\centering
\includegraphics[width=0.85\linewidth]{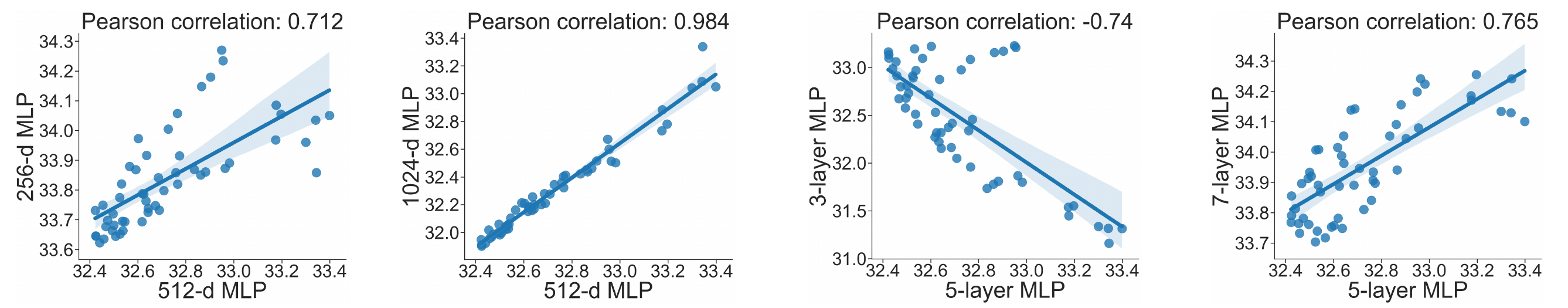}
\caption{
\textbf{Ablation study: varying width/depth of neural networks.}
Pearson correlation between MLPs with different width/depth over the same set of 52 relational graphs. 5-layer 512-d MLP is the default architecture.}
\label{fig:corr_widthdepth}
\end{figure*}

\section{Details for Matching the Reference FLOPS}

Here we provide more details on matching the reference FLOPS for a given model.
As described in Section 3.3, we vary the layer width of a neural network to match the reference FLOPS.

Specifically, to match the reference FLOPS, if all the layers have the same width, we incrementally vary the layer width by 1 for all the layers, then pick the layer width that has the fewest FLOPS above the reference FLOPS.
In the scenarios where layer width varies in different stages, we fine-tune the network width in each of the stages via an iterative mechanism: (1) we incrementally vary the layer width of the narrowest stage by 1, while maintaining the ratio of layer width across all the stages, then fix the layer width for that stage; (2) we repeat (1) for the narrowest stage in the remaining stages.
Using this technique, we can control the complexity of a model within 0.5\% of the baseline FLOPS.

\section{Details for Wall Clock Running Time}
Training a baseline MLP (translated by a complete graph) on CIFAR-10 roughly takes 5 minutes on a NVIDIA V100 GPU, while training all baseline ResNets and EfficientNets approximately take a day on 8 NVIDIA V100 GPU.
Due to the lack of mature support of sparse CUDA kernels, we implement relational graphs via applying sparse masks over dense weight matrices. The most sparse graph that we experiment with (sparsity = 0.125) is around 2x slower (in wall clock time) than the corresponding baseline graph.

\begin{figure*}[!t]
\centering
\includegraphics[width=0.92\linewidth]{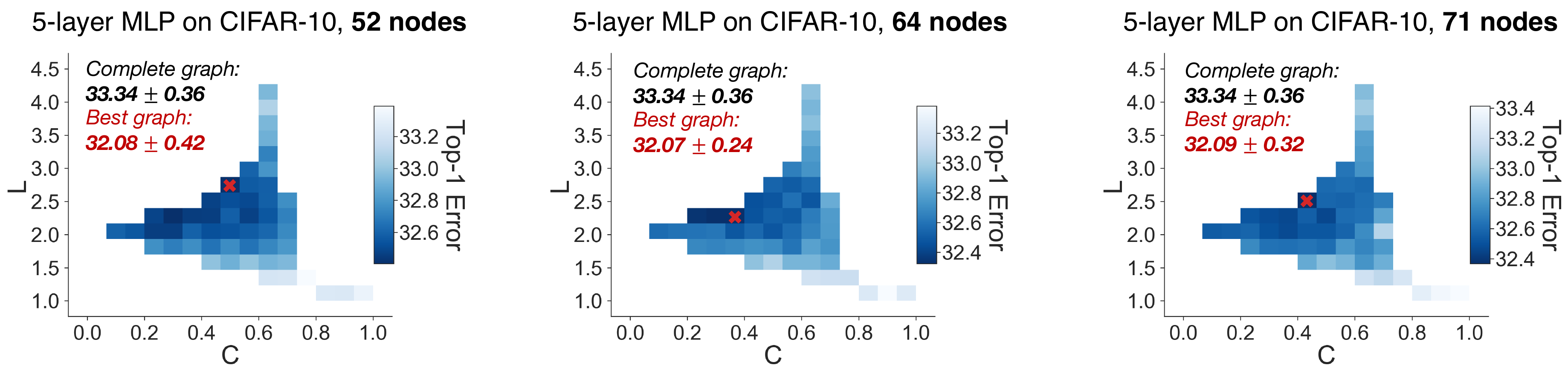}
\caption{
\textbf{Ablation study: varying number of nodes in relational graphs.}
We average results from 482 relational graphs for 52-node graphs, 449 graphs for 64-node graphs, and 422 graphs for 71-node graphs.
}
\label{fig:node}
\end{figure*}

\section{Analysis with More Graph Measures}
In the paper, we focus on 2 classic graph measures: clustering coefficient and average path length. Here we include the analysis using more graph measures.
Specifically, we consider the following additional local and global graph measures. 

\xhdr{Local graph measures}
(1) Average degree: node degree averaged over all the nodes. 
(2) Local efficiency: a measure of how well information is exchanged by a node's neighbors when the node is removed, averaged over all the nodes.

\xhdr{Global graph measures}
(1) Algebraic Connectivity: the second smallest eigenvalue of graph Laplacian. 
Graph Laplacian is defined as $A-D$, where $A$ is the adjacency matrix and $D$ is the degree matrix.
(2) Global efficiency: a measure of how well information is exchanged across the whole network.

We plot the performance of 5-layer MLPs on CIFAR-10 dataset versus one of the graph structure measures, over the 3942 relational graphs that we experimented with.
We use locally weighted linear regression to visualize the overall trend.
From Figure \ref{fig:1d_stats}, we can see that more graph measures exhibit the interesting U-shape correlation with respect to neural network predictive performance.

\section{Ablation Study}

\subsection{Varying Width/Depth of Neural Networks}
Here we investigate the effect of network width/depth on the performance of neural networks translated by the same set of relational graphs.
Specifically, we study 5-layer MLPs with $[256, 512, 1024]$ dimension hidden layers, and 512-dim MLP with $[3, 5, 7]$ layers. We can see that the performance of relational graphs with certain structural measures highly correlates across networks with different width.

The behavior is different when varying the network depth: while increasing the depth of MLP to 7 layers maintains a high correlation with 5-layer MLP, decreasing the depth of MLP to 3 layers completely reverse the correlation\footnote{Recall that when translating a relational graph, we leave the input and output layer unchanged; therefore, a 3-layer MLP only has 1 round of message passing over a given relational graph.}.
One possible explanation is that while sparse relational graphs may represent a more efficient neuron connectivity pattern, more rounds of message exchange are necessary for these neurons to fully communicate.
Understanding how many rounds of message exchange are required by a given relational graph to reach optimal performance is an interesting direction left for future work. 

\subsection{Varying Number of Nodes in Relational Graphs}
In Section 2.3, we show that an $m$-dim neural network layer can be flexibly represented by an $n$-node relational graph, as long as $n\leq m$. Here we show that varying the number of nodes in a relational graph has little effect on our findings.

In Figure \ref{fig:node}, we show the results of 5-layer MLP on CIFAR-10, where we consider using 52-node (number of nodes of the cat cortex graph) and 71-node (number of nodes of the macaque whole cortex graph) relational graphs in addition to 64-node graphs used in the main paper.
To cover the space of clustering $C$ and path length $L$, we generate 482 graphs for 52-node graphs, 449 graphs for 64-node graphs, and 422 graphs for 71-node graphs. To save computational cost, we use fewer graphs than the 3942 graphs in the main paper.
From the results, we can see the performance of the best graph is almost identical across these varied number of nodes, which justifies our claimed flexibility of selecting the number of nodes in a relational graph.

\begin{figure*}[!t]
\centering
\includegraphics[width=\linewidth]{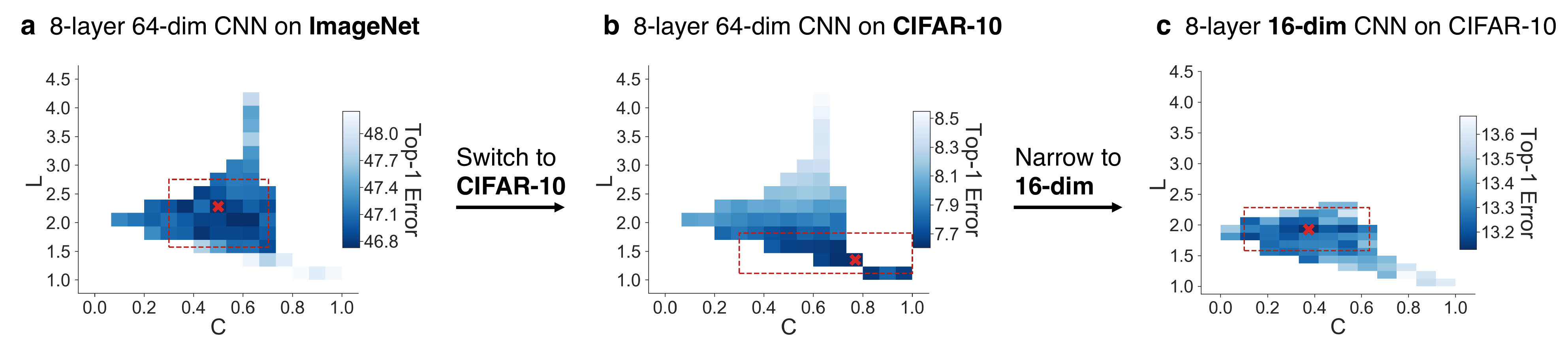}
\caption{
\textbf{(a)} We recap the results of 8-layer 64-dim CNN on ImageNet.
\textbf{(b)} We apply the same architecture to CIFAR-10. Results from 449 relational graphs are averaged to 52 bins, using the technique in Section 5.2. We find that the complete graph performs better than most sparse graphs in this setting.
\textbf{(c)} We hypothesize that since CIFAR-10 is an intrinsically much easier task than ImageNet, the graph structural priors might not matter when the representation capacity is abundant. To verify the hypothesis,we reduce the model capacity by reducing the width of CNN from 64 to 16 dimensions. 
Due to the narrowed dimensions, we explore relational graphs with 16 nodes instead of 64. Results of 326 relational graphs are averaged to 48 bins.
In this setting, sparse graph structure significantly outperforms the complete graph again.
}
\label{fig:overparam}
\end{figure*}

\begin{figure*}[!t]
\centering
\includegraphics[width=0.8\linewidth]{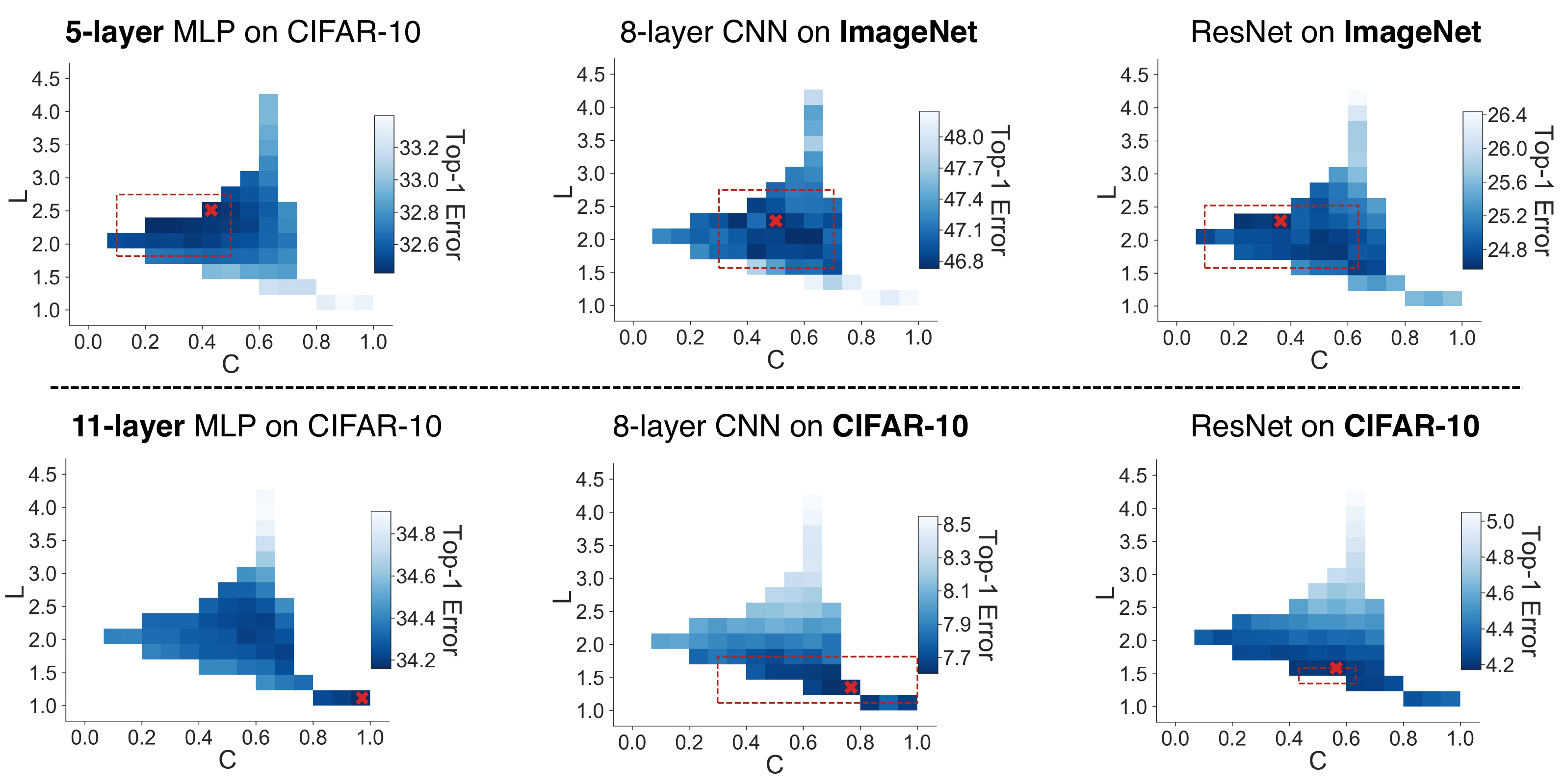}
\caption{
There are some special cases where the consistent sweet spots disappear, when training CNN models on CIFAR-10 (bottom row). For a simple task, we hypothesize that the graph structural priors might not matter when the representation capacity is abundant. However, for challenging tasks like ImageNet, sweet spots are consistent and sparse relational graphs always produce better results than the fully-connected/complete graph counterpart. (For 11-layer MLP on CIFAR-10, no sweet spot can be identified.)}
\label{fig:overparam2}
\end{figure*}

\section{Discussion of Failure Cases}
There are some special cases for convolutional neural networks on CIFAR-10, where the consistent sweet spot pattern ($C \in [0.43, 0.50]$, $L \in [1.82, 2.28]$) breaks and the best results are obtained with approximately fully-connected graphs. We visualize and discuss this phenomenon in Figure \ref{fig:overparam}. We start from analyzing the results of 8-layer 64-dim CNN on ImageNet, where consistent sweep spot region emerges (Figure \ref{fig:overparam}(a)). We then hold the model fixed, and switch the dataset to CIFAR-10. On CIFAR-10, we find that the complete graph performs better than most sparse graphs (Figure \ref{fig:overparam}(b)).

Without a theoretical discussion on network overparameterization~\cite{zhang2016understanding} and intrinsic task difficulty~\cite{li2018measuring}, we provide some empirical intuitions for those cases. As we have shown in main paper Figure 7,
fully-connected neural networks can automatically learn graph structures during training. We hypothesize that the structural prior matters less when the task is simple, relative to the representation capacity and efficiency of the neural network and the learned graph structure is sufficient in those settings.

To verify this hypothesis, we reduce the model capacity by reducing the width of CNN from 64 to 16 dimensions (Figure \ref{fig:overparam}(c)), and train it again on CIFAR-10. In this setting, we show that sparse graph structure significantly outperforms the complete graph again.
We provide more examples that compares two scenarios in Figure \ref{fig:overparam2}. Note that with ImageNet, sparse relational graphs consistently yield better performance in the sweet spot regions, even when the model capacity is increased.


\bibliography{bibli}
\bibliographystyle{icml2020}